\documentclass[10pt,leqno]{amsart}
\usepackage[T1]{fontenc}
\usepackage[utf8]{inputenc}
\usepackage{lmodern}
\usepackage{microtype}
\usepackage[a4paper,margin=2.5cm]{geometry}
\usepackage{amsmath,amssymb,amsthm}
\usepackage{mathtools}
\usepackage{graphicx}
\usepackage{float}      
\usepackage{booktabs}
\usepackage{tabularx}
\usepackage{csquotes}
\usepackage{cite}
\usepackage[obeyspaces]{url}
\usepackage[hidelinks]{hyperref}

\usepackage{titlesec}
\titleformat{\section}{\normalfont\Large\bfseries}{\thesection}{0.6em}{}
\titlespacing*{\section}{0pt}{2.0ex plus 0.5ex}{1.0ex plus 0.2ex}
\titleformat{\subsection}{\normalfont\large\bfseries}{\thesubsection}{0.6em}{}
\titlespacing*{\subsection}{0pt}{1.5ex plus 0.3ex}{0.6ex plus 0.2ex}
\titleformat{\subsubsection}{\normalfont\normalsize\bfseries}{\thesubsubsection}{0.6em}{}
\titlespacing*{\subsubsection}{0pt}{1.0ex plus 0.3ex}{0.4ex plus 0.2ex}

\begin{document}

\title{Predictive Modeling of Maritime Radar Data Using Transformer Architecture} 
\author{Bjorna Qesaraku, Jan Steckel}
\date{\today}
\address{Cosys-lab, Faculty of applied Engineering, University of Antwerp}
\email{jan.steckel@uantwerpen.be, bjorna.qesaraku@student.uantwerpen.be}
\maketitle

\let\thefootnote\relax

\begin{abstract}

Maritime autonomous systems require robust predictive capabilities to anticipate vessel motion and environmental dynamics. While transformer architectures have revolutionized AIS-based trajectory prediction and demonstrated feasibility for sonar frame forecasting, their application to maritime radar frame prediction remains unexplored, creating a critical gap given radar's all-weather reliability for navigation. This survey systematically reviews predictive modeling approaches relevant to maritime radar, with emphasis on transformer architectures for spatiotemporal sequence forecasting, where existing representative methods are analyzed according to data type, architecture, and prediction horizon. Our review shows that, while the literature has demonstrated transformer-based frame prediction for sonar sensing, no prior work addresses transformer-based maritime radar frame prediction, thereby defining a clear research gap and motivating a concrete research direction for future work in this area.
\end{abstract} 

\section{Introduction}

The shipping industry has historically been conservative in adopting technological innovations. However, the recent technical innovations have increased the pressure to develop more capable perception systems, as autonomous and semi-autonomous vessels transition from research concepts to real operations. Recent surveys by Qiao et al.~\cite{10021250} and Thombre et al.~\cite{Thombre2020} show that artificial intelligence, especially deep learning, has become central to maritime situational awareness, excelling in tasks such as object detection, collision risk assessment, and trajectory prediction. This shift toward autonomous vessels is driven by both safety and economic factors, as human error contributes to roughly 58 percent of maritime casualties in European waterways alone in the last decade~\cite{EMSA2024}. At the same time, maritime transport remains the core of global commerce, carrying over 80 percent of world trade by volume, so even modest improvements in safety and efficiency can have a great economic impact~\cite{UNCTAD2024}. Therefore, autonomous navigation systems are expected to also support more resilient and reliable shipping in this increasingly complex maritime environment, which requires perception systems that do more than merely describe the current scene: they must anticipate how the maritime environment will evolve over time.

\begin{figure}[H]
    \centering
    \includegraphics[width=\linewidth]{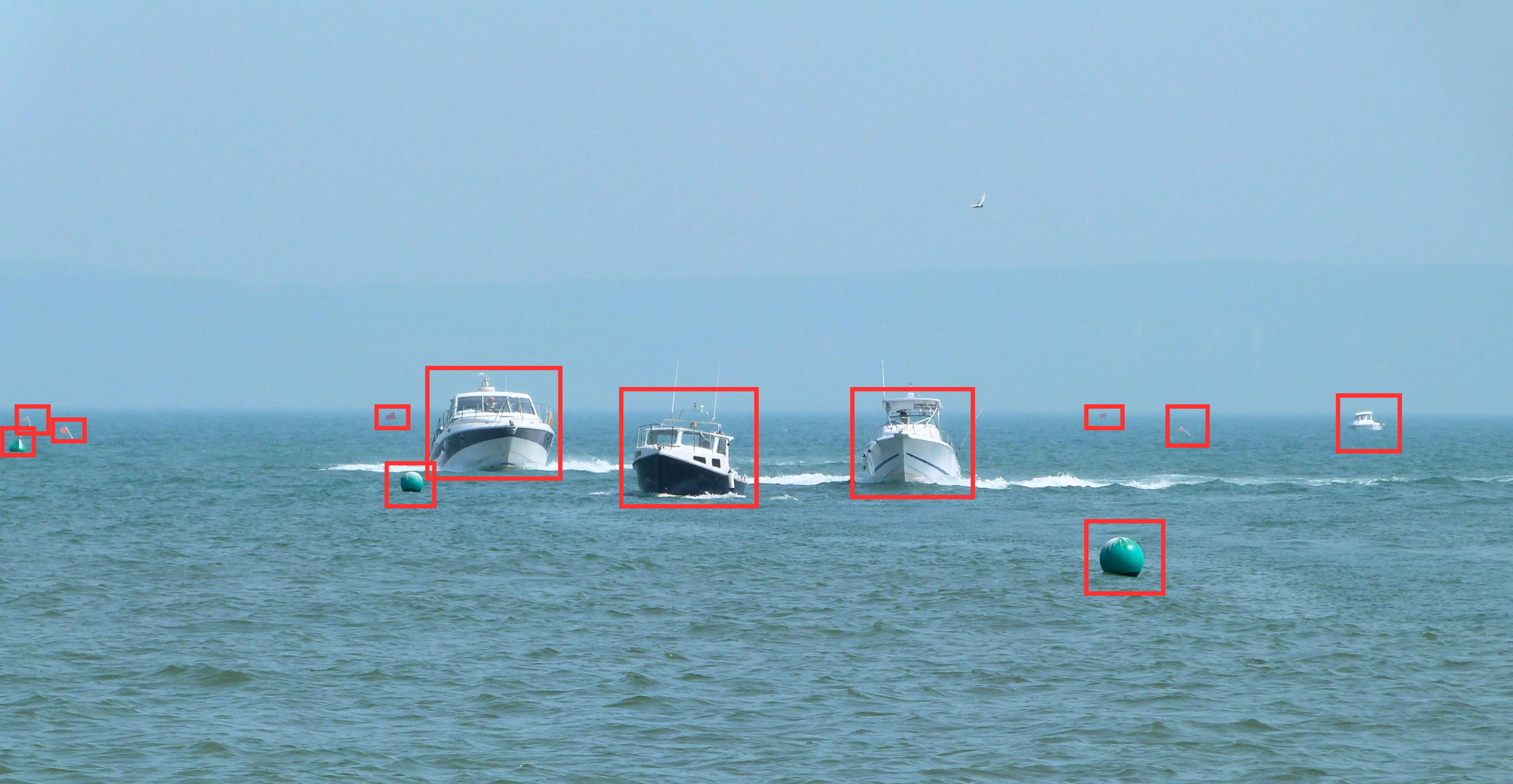}
    \caption{Annotated maritime scene showing multiple objects (small vessels, distant ships and navigation buoys and flags) that must be detected and classified by the perception system. Base photograph by Charlotte Clark.}
    \label{fig:maritime_scene}
\end{figure}

Maritime radar data is characterized by a set of properties that make it both attractive and challenging for predictive modelling. Compared with optical cameras and LiDAR, radar remains reliable in fog, rain and rough seas, which makes it a key all-weather sensing modality for navigation and collision avoidance in challenging conditions~\cite{Ersu2024}. In particular, X-band marine radars can cover several kilometers with typical update periods of about 1 to 2 seconds and spatial resolutions on the order of 5 to 10 meters, providing dense temporal sampling over a wide area and therefore supporting long-horizon forecasting~\cite{Neill2018}. Nevertheless, marine radar images are sparse and heavily contaminated by sea clutter, leading to missed object detections as well as false alarms. Furthermore, the peculiar reflection regime of radar causes a signal structure that differs significantly from natural images on which many deep learning models are originally trained~\cite{Zhu2020}. These peculiarities are further shaped by the complexity of the maritime scene itself, where vessels interact with other dynamic actors and static coastlines or ports continuously. Figure~\ref{fig:maritime_scene} illustrates a maritime scene containing multiple vessel types, navigation aids and buoys at varying ranges, that perception systems must simultaneously monitor and predict.

Recent surveys have examined maritime prediction from complementary perspectives. Concretely, Li et al.~\cite{Li2023} analyzed 64 unique trajectory prediction approaches, where 28 of the reviewed methods apply deep learning methods. In contrast, Xie et al.~\cite{Xie2025} review 16 transformer-based models used for mixed short-term and long-term trajectory prediction, where the source of data comes mostly from the Automatic Identification System (AIS), which is a maritime tracking and anti-collision tool that broadcasts vessel details such as identification, position, and course. Geng et al.~\cite{Geng2021} survey machine learning in radar signal processing, focusing on classification and clutter suppression. However, as illustrated in Figure~\ref{fig:maritime_scene}, maritime scenes contain far more information than discrete vessel coordinates, yet no existing survey addresses the raw frame-level prediction for maritime radar. This represents a critical gap since autonomous systems require the capabilities of anticipating the evolution of the full scene and not just vessel trajectories. This survey aims to bridge this gap by providing the first systematic analysis of predictive modeling approaches for maritime radar imagery, examining the potential of transformer architectures for spatiotemporal radar sequence forecasting, and identifying specific technical challenges and research opportunities for radar frame prediction.
 
\section{Background}

\subsection{Maritime Radar Fundamentals}

Maritime radar systems create two-dimensional images by emitting electromagnetic pulses and measuring the time delay, intensity, and frequency shift of reflected signals from objects in the environment. Through the analysis of the received signals, the radar systems can infer the bearing (position of antenna during scan rotation), the range (time-delay) and the radial velocity component (Doppler-shift) of the reflecting objects. Therefore, understanding the fundamental signal processing techniques of radars is essential for estimating the capabilities and challenges of radar frame prediction.

Range estimation is a technique whose fundamental principle lies in measuring the round-trip time $\tau$ between pulse transmission and echo reception, which determines the range $R$ to reflecting objects according to $R=c \tau/2$, where $c$ is the speed of light and the division by two accounts for the two-way propagation path. For a simple unmodulated pulse of duration $T$, the range resolution $\Delta R=cT/2$ represents the minimum separation between two targets that can be distinguished. However, short term pulses contain limited energy, which restricts the maximum detection range and requires high peak transmission power. In order to overcome the limitation of energy-resolution, modern radars employ pulse compression techniques that can achieve more precise detection over long distances by modulating the pulse frequency and applying matched filtering at reception~\cite{Richards2022}. The separation (i.e., the Rayleigh limit) will then be a function of the bandwidth of the received signal  ~\cite{VanTrees1968_DEMT1, VanTrees1971_DEMT2}.

When a radar and a target are not at rest with respect to each other, the frequency $F_r$ of the received echo will differ from the transmitted frequency $F_t$ due to the Doppler effect. Doppler estimation in radar systems is the process of measuring this frequency shift to detect whether the target is approaching the radar (i.e., it has a slightly higher frequency) or moving away from it (i.e., it has a lower frequency), and to determine the component of velocity along the line of sight, known as the radial velocity~\cite{Richards2022}.

Bearing estimation in radar systems refers to the process of determining the azimuth angle of a target relative to the radar's reference direction, such as true North or the direction of the ship. This is achieved by measuring the angle in the horizontal plane when the radar beam intercepts the target. The accuracy of bearing measurement is influenced by factors such as antenna beam width, signal processing resolution, and the presence of clutter or interference. Mechanically rotating antennas, which are standard in maritime applications, achieve narrow beam widths through large physical apertures, but are limited to rotation rates of 20--40~RPM to allow sufficient dwell time on each bearing for an adequate signal-to-noise ratio. This yields frame rates of 0.3--0.7~Hz for $360^\circ$ coverage or 1--2~Hz for sector scanning covering limited angular ranges~\cite{Bole2014}.

For predictive modeling, these fundamental radar-signal processing methods impose advantages and challenges simultaneously. On the positive side, pulse compression offers high range resolution that supports detailed spatial representations of vessels and coastlines, enabling predictive models to learn from rich patterns. Moreover, Doppler estimation supplies explicit velocity data, allowing models to integrate motion vectors rather than calculating speed from position changes, and bearing estimation facilitates convolutional or patch-token approaches by imposing a regular spatial grid structure, typical in vision-transformer frameworks.

However, several constraints apply, e.g. the mechanical scanning of X-band radar requires three seconds to complete a full $360^\circ$ rotation, resulting in a sensor update rate of approximately 0.3~Hz~\cite{Jang2024}. This relatively low frame rate creates significant temporal gaps between observations, which can disrupt continuous motion tracking and pose a challenge for smooth motion modeling. Furthermore, pulse-compression side lobes and Doppler blind speeds introduce ambiguous or missing features that predictive models must learn to handle. These challenges simply mean that model architectures must be tailored explicitly for radar data rather than directly importing those developed for natural-image or video domains.

\subsection{Transformer Architectures for Prediction}

Transformers~\cite{Vaswani2017}, originally developed for natural language processing, have revolutionized sequence modeling through self-attention mechanisms that capture long-range dependencies without recurrence. Given an input sequence $X$, the self-attention computes:
\begin{equation}
 \mathrm{Attention}(Q, K, V) = \mathrm{softmax}\left(\frac{QK^{\top}}{\sqrt{d_k}}\right) V 
\end{equation} 
where $Q$, $K$, $V$ are learned projections named queries, keys, and values respectively, and $d_k$ is the key dimension~\cite{Vaswani2017}. This enables each element to attend to all others, learning complex spatiotemporal relationships.

Vision transformers (ViT)~\cite{Dosovitskiy2021} adapt this modeling technique to images via patch tokenization by dividing images into non-overlapping patches and projecting them to embedding vectors. For video or sequence prediction, temporal transformers process sequences of frame embeddings, while spatio-temporal variants combine spatial and temporal attention~\cite{Bertasius2021}. What makes transformers particularly well-suited for radar frame predictions is the ability of self-attention to capture long-range spatio-temporal dependencies across several past frames, enabling more accurate forecasts over extended horizons. In addition, processing all temporal tokens in parallel during training removes the sequential limitations of recurrent networks and makes it easier to use long radar histories and larger batch sizes effectively. Finally, the demonstrated success in sonar frame prediction using transformers in the EchoPT paper~\cite{Steckel2024} suggests a natural path to the radar sensing modality.

\subsection{Advantages of Predicting Radar Frames}

\begin{figure}[H]
    \centering
    \includegraphics[width=\linewidth]{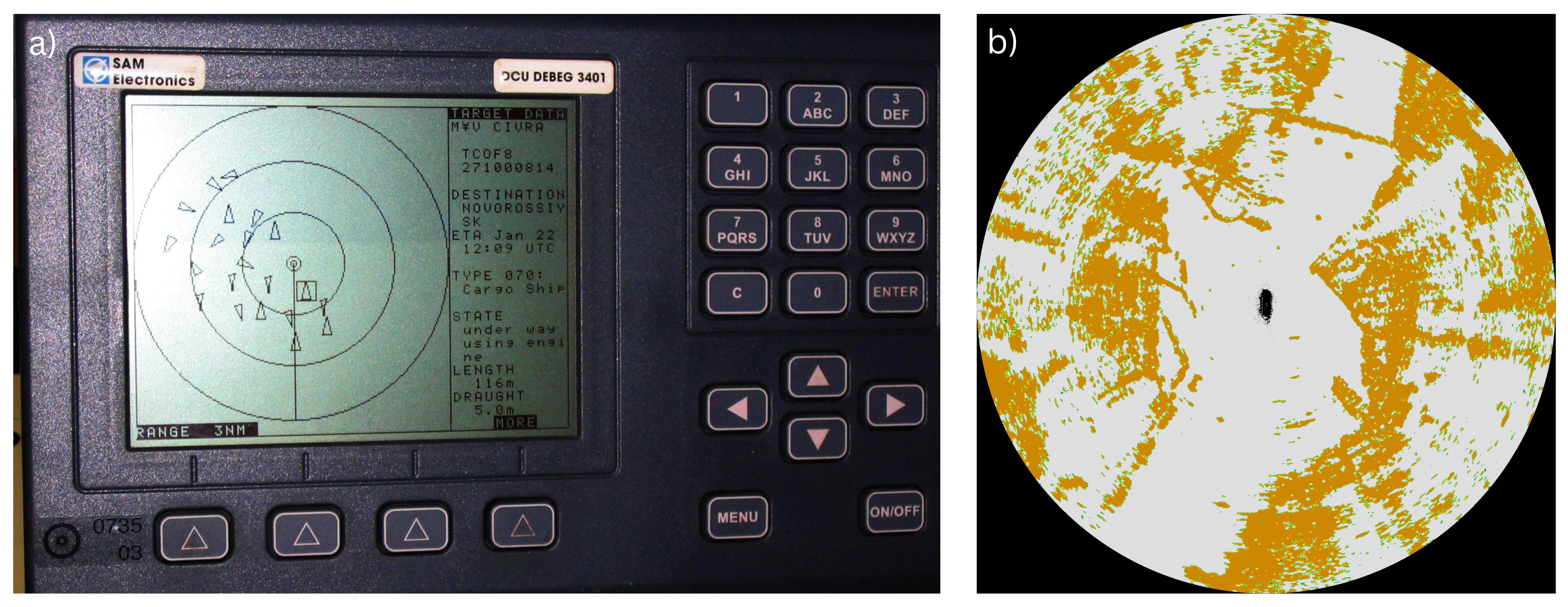}
    \caption{Comparison of AIS and radar observations. (\textbf{a}) AIS display/control unit shows only cooperative vessels that can broadcast AIS messages like identity, voyage data, etc. Image by Clipper, ``Ais dcu bridge'', 2006, Wikimedia Commons, licensed under CC BY 2.5~\cite{AIS2006}. (\textbf{b}) Example of an X-band radar frame from MOANA dataset~\cite{Jang2024} containing the full radar scene, including shoreline returns and additional small echoes that may correspond to clutter, navigation aids, or buoys.}
    \label{fig:aisradar}
\end{figure}

Existing maritime transformer models forecast discrete vessel positions from AIS or tracking data, achieving impressive long-horizon accuracy~\cite{Nguyen2021,Li2023}. However, these coordinate-based models operate on sparse data that discard the rich spatial information present in radar observations, as shown in Figure~\ref{fig:aisradar}. Alternatively, forecasting entire future radar images addresses critical perception capabilities that trajectory methods cannot provide. First, frame prediction preserves complete spatial context by generating 2D images that encode not only vessel positions but also coastlines, navigation aids, buoys, and small boats that are exempt from carrying AIS. This enhances the reasoning about safe passages and multi-vessel configurations that are essential for collision avoidance in crowded waterways. 

Second, unlike trajectory methods that require vessels to be detected and tracked before prediction, frame prediction operates at the sensor observation level, providing advance warning in congested environments where new objects continuously appear. Moreover, as vessels maneuver, their radar returns change with viewing angle~\cite{Richards2022}, so the frame prediction naturally captures these dynamics, supporting robust object tracking and classification even through maneuvers.

Finally, when detection fails due to heavy clutter or sensing equipment issues, trajectory methods lose their predictive capability, whereas frame-prediction methods can maintain situational awareness by generating expected radar observations. Therefore, calculating the differences between the expected frame and the observed frame enables sensing equipment anomaly detection at an early stage. In addition, frame prediction supports collision avoidance over a short horizon, while trajectory prediction methods can forecast minutes to hours ahead, making them more helpful in strategic route planning. To conclude, this complementary approach shows that frame prediction addresses fundamental capabilities that are essential for robust maritime autonomy, which coordinate-based methods alone cannot provide.

\section{Literature Review}

\subsection{Traditional and Machine Learning Approaches in Maritime Applications}

\subsubsection{Traditional Methods}

Early maritime prediction has primarily been used to predict the vessel trajectory and has relied on hand-crafted physics-based motion models with Gaussian noise assumptions. Perera and Guedes Soares~\cite{Perera2010} design a kinematic ship model and apply the Extended Kalman Filter to fuse radar or AIS data, enabling real-time state estimation and short-term trajectory prediction for collision avoidance. Moreover, Rong et al.~\cite{Rong2019} train a Gaussian model on AIS tracks to produce probabilistic forecasts of future positions using mean and variance. These models are lightweight, interpretable, and easy to integrate into existing navigation systems, however, their simplified dynamics make them operate poorly under abrupt maneuvers, multi-vessel interaction, and in complex maritime environments. 

\subsubsection{Machine Learning (ML) Methods}

The predictive models that used machine learning approaches have relied extensively on AIS data. Liu et al. have proposed an ACDE-SVR model that combines adaptive chaos differential evolution with support vector regression to forecast vessel position, speed and heading from short AIS histories, achieving higher and more stable accuracy at low computational cost~\cite{Liu2019}. Furthermore, Zhang et al. build a random forest model on large-scale AIS data to predict vessel destinations by measuring the similarity between current and historical trajectories, reporting high classification accuracy in busy traffic waterways~\cite{Zhang2020}. Even though these models have succeeded in capturing nonlinear patterns more accurately than traditional methods, they have still relied on manually engineered features, which was further improved with deep learning.

\subsection{Deep Learning (DL) Approaches in Maritime Applications}

\subsubsection{CNN for Target Detection}

Convolutional Neural Networks (CNN) excel at spatial feature extraction from radar imagery. For example, Chen et al.~\cite{Chen2021} fuse multi-domain features via CNNs for small target detection, and Qu et al.~\cite{Qu2021} employ an attention-enhanced CNN to capture and learn the deep features of Wigner--Ville distribution of the radar signals. However, pure CNNs lack temporal modeling since they process single frames without capturing motion dynamics that are necessary for prediction.

\subsubsection{RNNs for Trajectory Prediction}

Recurrent Neural Networks (RNNs) model temporal dependencies which make them suitable for sequence prediction. Wan et al.~\cite{Wan2019} propose Bi-LSTM using instantaneous phase, Doppler spectrum, and STFT features for sea clutter target detection, achieving an average detection probability of the sequence-feature detector of 0.955. For AIS trajectory prediction, Li et al.~\cite{Li2023} review 28 DL approaches, predominantly Long Short Term Memory (LSTM) variants, which have a high success rate in forecasting the path. However, when it comes to frame-level prediction, RNNs struggle with long-range dependencies and have difficulties in capturing spatial relationships within 2D frames. 

\subsection{Transformer-Based Approaches for Next-Frame Prediction}

\subsubsection{AIS Trajectory Prediction}

AIS data provides precise vessel coordinates, enabling trajectory forecasting as a sequence-to-sequence problem. Zhou et al.~\cite{Nguyen2021} have proposed a transformer specifically for vessel trajectory prediction from AIS sequences. In their architecture, they have incorporated latitude, longitude, speed, and course in positional encoding, multi-head self-attention (8 heads) capturing vessel interaction patterns, temporal encoding via sinusoidal functions, and a decoder generating future positions autoregressively. They achieve 10-minute ADE (Average Displacement Error) of 145~m vs. 267~m for LSTM baselines, showing a 45\% improvement. The model handles up to 60-minute predictions but degrades beyond 30 minutes due to accumulating errors. Its key limitation is that it operates on sparse coordinate sequences, not dense spatial 2D radar imagery. Furthermore, in their survey, Xie et al.~\cite{Xie2025} have analyzed 16 transformer variants for maritime tasks and have observed this common pattern: transformers consistently outperform RNNs for long-horizon trajectory prediction ($>$15 min) due to superior long-range dependency modeling; however, they all work with coordinate sequences rather than raw sensor imagery.

\subsubsection{Object Detection with Transformers}

While transformers have not been applied to radar frame prediction yet, they appear in target detection pipelines. He et al.~\cite{He2025} integrate transformer blocks into YOLOv5 for improved object detection. Their transformer mechanism patches multi-frame radar images into tokens, applies self-attention to learn inter-frame relationships, and enhances features fed to YOLO detector. Their architecture uses 6 transformer layers with 8 attention heads and their results prove that transformers can process radar imagery effectively. They are able to detect objects in these radar frames as they move across frames, however, their system outputs bounding boxes encapsulating the objects. Extending the use of transformers to full radar frame prediction remains unexplored yet.

\subsubsection{Sonar Frame Prediction - EchoPT}

This is the most relevant prior work that demonstrates frame-level prediction feasibility for acoustic imaging. Steckel et al.~\cite{Steckel2024} propose a transformer-based model capable of predicting 2D sonar images from historical frames and ego-motion. Their model uses individual sonar frames that are divided into $P \times P$ patches (with $P=16$), linearized, projected into 768-dimensional embeddings, augmented with learned positional encodings, and concatenated with separately embedded ego-motion of the robot (velocity and angular rate). A transformer encoder of 12 layers and 12 heads processes the sequence of past sonar frames $(F_{t-k}, \dots, F_{t-1})$, and a linear decoder maps the output tokens back to the pixel space. The model is pre-trained on large-scale mobile-robot sonar data using a combination of MSE (Mean Squared Error) and perceptual losses, achieving high structural similarity for short prediction horizons and gradually degrading as error accumulates over long prediction horizons. To conclude, EchoPT shows that patch tokenization, explicit ego-motion conditioning, and perceptual loss terms are effective for sparse 2D sensing, however, it operates on dense, short-range, sonar images at higher frame rates. In contrast, maritime radar is slower, long-range and with different clutter dynamics, so adapting this architecture to radar frames requires handling larger temporal gaps, sea clutter, and larger-scale spatial structure.

\begin{table} 
\caption{Representative methods for maritime perception and prediction.\label{tab:maritime_methods}}
\begin{tabularx}{\textwidth}{XXXXX}
\toprule
\textbf{Method} & \textbf{Data} & \textbf{Task} & \textbf{Architecture} & \textbf{Horizon}\\
\midrule
Perera~\cite{Perera2010} & Radar/AIS & Trajectory prediction & EKF & $\sim$min \\
Rong~\cite{Rong2019} & AIS & Trajectory prediction & GP & 30--60 min \\
Liu~\cite{Liu2019} & AIS & Trajectory prediction & ACDE-SVR & $\le$30 min \\
Zhang~\cite{Zhang2020} & AIS & Destination prediction & RF & Voyage \\
Chen~\cite{Chen2021} & Radar & Object Detection & CNN & Real-time \\
Qu~\cite{Qu2021} & Radar & Object detection & Att.-CNN & Real-time \\
Wan~\cite{Wan2019} & Radar & Object detection & Bi-LSTM & Real-time \\
TrAISformer~\cite{Nguyen2021} & AIS & Trajectory prediction & Transformer & 1--3 h \\
DTNet~\cite{He2025} & Radar & Detection and tracking & CNN+Transf. & Real-time \\
EchoPT~\cite{Steckel2024} & Sonar & Frame prediction & Transformer & Few steps \\
\bottomrule
\end{tabularx}
\end{table}

\subsection{Video Prediction Models}

Transformer-based video prediction in computer vision offers useful templates for radar frame forecasting. VideoGPT~\cite{Yan2021} models videos in two stages: first, a 3D-convolutional VQ-VAE first compresses the video into discrete spatio-temporal latents, and then a GPT-style transformer autoregressively predicts future latent tokens with positional encodings, achieving competitive Fréchet Video Distance on datasets like BAIR and UCF-101. Alternatively, MaskViT~\cite{Gupta2023} pre-trains transformers by masked visual modeling, using separate spatial and spatio-temporal window attention and iterative token refinement to generate high-resolution (256$\times$256) future frames more efficiently than prior models, while supporting goal-conditioned prediction. Moreover, VPTR~\cite{Ye2023} introduces an efficient local spatio-temporal attention block and compares fully autoregressive, partially autoregressive, and non-autoregressive transformer variants, showing performance competitive with ConvLSTM baselines on standard video prediction benchmarks.

These designs are quite relevant to maritime radar frame prediction, e.g., latent tokenization, as in VideoGPT, allows transformers to model large, noisy radar frames in a compact representation, reducing computational cost and focusing on high-level structure. Furthermore, efficient attention mechanisms, as explored in VPTR, enable scalable modeling of long radar sequences by restricting attention to local spatiotemporal neighborhoods. In addition, masked video pre-training, as in MaskViT, offers a way to exploit large volumes of unlabeled radar data, improving sample efficiency and helping the model learn generic clutter and coastline structure before supervised frame prediction.

\section{Discussion and Future Directions}

Table~\ref{tab:maritime_methods} provides a structured representation of the key methods covered in this survey, organized by perception modality, tasks such as trajectory prediction, detection and tracking, and architectures used to facilitate these tasks. For predictive approaches, the horizon of forecasting is also mentioned in a separate column. From the table, it is apparent that AIS is used almost exclusively for long-horizon trajectory or destination prediction, while radar is used for real-time object detection and tracking, with only one traditional EKF approach combining both. However, even though transformers have already been adopted for maritime tasks, there is, to date, no existing work on transformer-based prediction of future maritime radar frames, besides related formulations explored for optical video and sonar imagery~\cite{Steckel2024}.

Existing studies only partially address radar-specific challenges such as low frame rates, dynamic and state-dependent sea clutter, and polar sampling with range-dependent resolution. In addition, public datasets suitable for prediction are scarce, and most available maritime radar data focus on single-frame object detection rather than multi-frame sequences with reliable ground truth. Evaluation practice is also fragmented, since frame prediction is typically assessed with generic image metrics, while coordinate-based methods report trajectory errors that cannot be directly compared to frame-level metrics.

Despite these limitations, frame-level prediction offers capabilities that coordinate-only approaches cannot provide. Predicting future radar scenes would enable systems to reason about vessels that are not yet confidently tracked, anticipate changes as vessels maneuver, and model the evolution of clutter and other environmental conditions. Such capabilities are directly relevant to collision avoidance, anomaly detection, and anticipatory perception under degraded sensing. Additionally, the availability of the MOANA dataset~\cite{Jang2024} is a crucial enabler in this context, since it provides large-scale, multi-band maritime radar sequences with associated multimodal information, making it possible to train and systematically evaluate data-intensive deep models for frame prediction under diverse conditions and radar configurations.

Within this context, several focused research directions emerge. A natural first step is to adapt the EchoPT transformer architecture from sonar to maritime radar, with appropriate modifications to address the specific characteristics of radar-perceived images. Another approach could be the exploration of hybrid methods that combine coordinate-based trajectory prediction with frame-level forecasting. This could exploit complementary strengths such as using trajectories for long-horizon prediction and frames for short- to medium-horizon spatial detail, potentially linked via cross-attention between coordinate and image embeddings. Furthermore, self-supervised pre-training on large volumes of unlabeled radar data offers a promising route to improve data efficiency and generalization across radar bands, platforms, and geographic regions. Finally, real-time deployment on vessel hardware will require making these models more efficient, for example, through knowledge distillation, pruning, quantization, or alternative attention and spatio-temporal architectures. Together, these techniques define a concrete and technically achievable agenda for future work on maritime radar frame prediction.

As we have argued before, Radar data, by virtue of its predominately specular reflection regime, is physically a very different sensing modality compared to optical techniques. Indeed, the relatively long wavelengths render the world virtually flat compared to the wavelength, promoting specular reflections over diffuse reflections. Furthermore, the direct possibility to sense the radial velocity component of the detected objects differentiate it even further from camera or LiDAR-based techniques. Therefore, we advocate that machine learning models for radar (and sonar, for that matter) take these physical properties explicitly into account. There are several recent examples where physically informed machine-learning models show great benefit in radar ~\cite{Zheng2024Redefining} and sonar ~\cite{Jansen2024SonoNERFs}. These physical properties that should be taken into account include the range/Doppler/bearing ambiguity functions. Indeed, there exists a tradeoff between the resolutions in range, bearing, and radial velocity through a concept very similar to the Heisenberg uncertainty principle ~\cite{Klemm2006STAP}. Additionaly, as radar data is most-naturally represented in a spherical coordinate system, vessel motion translates non-linearly into this coordinate system. This has been shown extensively for sonar sensing ~\cite{SteckelPeremans2017_AcousticFlowControl, PeremansSteckel2014_AcousticFlowICRA,JansenLaurijssenSteckel2022_AdaptiveAcousticFlow}, but we hypothesize that this is of equal importance for maritime radar data.

\section{Conclusions}

This survey has examined predictive modeling for maritime radar data with particular emphasis on transformer architectures and frame-level forecasting. Using a comparative analysis, we contrasted traditional signal processing, classical machine learning, and deep learning approaches across AIS-based trajectory prediction, radar-based detection and tracking, and frame prediction in closely related domains. The review shows that transformers are now well-established for AIS trajectory forecasting (e.g., TrAISformer~\cite{Nguyen2021}) and have demonstrated feasibility for sonar frame prediction (EchoPT~\cite{Steckel2024}), whereas their application to maritime radar imagery for future frame prediction remains unexplored.

A central conclusion is that frame-level radar prediction offers capabilities that go beyond coordinate-based forecasting. Trajectory models efficiently predict vessel positions over long horizons in a compact state space, but they do not capture the full spatial structure of the scene, the behavior of untracked objects, or the evolution of clutter and environmental conditions. In contrast, predicting complete future radar frames can provide spatial context for collision avoidance, enable anomaly detection through discrepancies between predicted and observed scenes, and maintain robust perception under degraded sensing, making trajectory-based and frame-based approaches complementary rather than competing.

These findings indicate that transformer-based maritime radar frame prediction is both technically tractable and practically relevant. The availability of large-scale datasets such as MOANA~\cite{Jang2024}, the demonstrated success of transformers in AIS and sonar applications, and the suitability of self-supervised learning for exploiting unlabeled radar streams create favorable conditions for progress. Future work that tailors transformer architectures to radar-specific characteristics and addresses computational efficiency for on-board deployment can fill the identified research gap and advance anticipatory maritime perception in support of safety and autonomy at sea navigation.

\vspace{6pt} 


\end{document}